\documentclass[11pt]{article}

\usepackage[]{emnlp2022}

\usepackage{times}
\usepackage{latexsym}
\usepackage{tabularx}

\usepackage{tikz}
\usetikzlibrary{decorations.pathreplacing,calc}
\newcommand{\tikzmark}[1]{\tikz[overlay,remember picture] \node (#1) {};}

\usepackage[english]{babel}
\usepackage{amsmath,amssymb, amsthm}
\usepackage{pgfplots}
\usepackage{algorithm}
\usepackage{algpseudocode}
\usepackage{varwidth}%
\usepackage{arydshln}

\DeclareMathOperator*{\argmax}{arg\,max}

\DeclareMathOperator*{\topk}{top-k_1}
\DeclareMathOperator*{\topkk}{top-k_2}

\usepackage{tabularx}
\usepackage{booktabs}
\usepackage{enumitem}
\usepackage{graphicx}
\usepackage{multirow}
\usepackage{float}
\usepackage[T1]{fontenc}
\usepackage{amsmath}
\usepackage{mathtools}

\usepackage[utf8]{inputenc}
\usepackage{cleveref}
\usepackage{csquotes}

\usepackage{microtype}
\usepackage{inconsolata}
\newcommand{\linebreakcell}[2][c]{%
  \begin{tabular}[#1]{@{}c@{}}#2\end{tabular}}
\newcommand{\tabincell}[2]{\begin{tabular}{@{}#1@{}}#2\end{tabular}}

\newcommand\blfootnote[1]{%
  \begingroup
  \renewcommand\thefootnote{}\footnote{#1}%
  \addtocounter{footnote}{-1}%
  \endgroup
}

\newcommand*{\textBF}[1]{\rlap{\raisebox{0pt}[0pt][0pt]{\textbf{#1}}}\phantom{#1}}

\makeatletter
\def\ALG@step%
   {%
   \addtocounter{ALG@line}{1}%
   \addtocounter{ALG@rem}{1}%
   \ifthenelse{\equal{\arabic{ALG@line}}{5}}%
      {\setcounter{ALG@rem}{0}\alglinenumber{\large\texttt{\textcolor{brown}{-}}}}%
      {\ifthenelse{\equal{\arabic{ALG@line}}{6}}%
      {\setcounter{ALG@rem}{0}\alglinenumber{\large\texttt{-}}}%
      {\ifthenelse{\equal{\arabic{ALG@line}}{7}}%
      {\setcounter{ALG@rem}{0}\alglinenumber{\large\texttt{-}}}%
      {\ifthenelse{\equal{\arabic{ALG@line}}{8}}%
      {\setcounter{ALG@rem}{0}\alglinenumber{\large\texttt{-}}}%
      {\ifthenelse{\equal{\arabic{ALG@line}}{9}}%
      {\setcounter{ALG@rem}{0}\alglinenumber{\large\texttt{-}}}%
      {\ifthenelse{\equal{\arabic{ALG@line}}{10}}%
      {\setcounter{ALG@rem}{0}\alglinenumber{\large\texttt{\textcolor{teal}{+}}}}%
      {}%
   }%
   }%
   }%
   }%
   }%
   }%
\makeatother
\algrenewcommand\alglinenumber[1]{\footnotesize #1}
\title{Controllable Factuality in Document-Grounded Dialog Systems Using a Noisy Channel Model}

\author{\bf Nico Daheim$^{1,*}$, David Thulke$^{2,3}$, Christian Dugast$^{3}$,
{\bf Hermann Ney$^{2,3}$ } \\
$^1$Ubiquitous Knowledge Processing Lab \\ Department of Computer Science, Technical University of Darmstadt \\
$^2$Chair for Human Language Technology and Pattern Recognition, RWTH Aachen University  \\
$^3$AppTek GmbH \\
{\url{www.ukp.tu-darmstadt.de}} \\
    \{thulke, ney\}@i6.informatik.rwth-aachen.de \\
}

\begin{document}
\maketitle
\begin{abstract}

In this work, we present a model for document-grounded response generation in dialog that is decomposed into two components according to Bayes' theorem.
One component is a traditional ungrounded response generation model and the other component models the reconstruction of the grounding document based on the dialog context and generated response.
We propose different approximate decoding schemes and evaluate our approach on multiple open-domain and task-oriented document-grounded dialog datasets.
Our experiments show that the model is more factual in terms of automatic factuality metrics than the baseline model.
Furthermore, we outline how introducing scaling factors between the components allows for controlling the tradeoff between factuality and fluency in the model output.
Finally, we compare our approach to a recently proposed method to control factuality in grounded dialog, CTRL \cite{rashkin2021ctrl}, and show that both approaches can be combined to achieve additional improvements.
\end{abstract}

\blfootnote{Code available at: \\ \url{https://github.com/ndaheim/noisy_channel_model}}
\blfootnote{* Work done while at RWTH Aachen University, Chair for Human Language Technology and Pattern Recognition}
  
\section{Introduction}
Recently, document-grounded dialog systems have seen an increase in popularity. %
Not only are they used to extend task-oriented systems beyond the narrow scope of fixed databases and APIs \cite{feng-etal-2020-doc2dial, kim-etal-2020-beyond, kim2021howrobustru, lamda2022}, but also to ground open-domain conversations in information from the web \cite{cmu_dog_emnlp18, dinan2018wizard, 2021internetaugmented} or in persona descriptions to make dialog agents more interesting and engaging \cite{zhang-etal-2018-personalizing}.
In any of these applications, the conversational system has to use the information from the document and blend it into the response \cite{roller2020opendomain}.
This means that the response should remain fluent, grammatically correct and coherent to the previous utterances in the dialog on the one hand, and on the other hand that it correctly reflects the information from the document.
This entails that no information is altered and also that no new information should be added if it is not immediately verifiable.
This is especially crucial in all cases where a user uses a system to satisfy an information need \cite{romewasbuiltin1776}.

While previous work has shown that retrieving relevant information is a crucial step for task-oriented \cite{kim-etal-2020-beyond} and open-domain document-grounded dialog systems and a potential mitigator of inconsistencies \cite{shuster-etal-2021-retrieval-augmentation}, there is sufficient evidence that grounded response generation models may still fail to produce factual responses, even when the correct information is contained in its grounding document.
In general, generating outputs that are both fluent and correct remains an open problem not only in dialog systems but natural language generation as a whole \cite{cao2018faithful, maynez2020faithfulness, roller2021recipes, 2021hallucinationSurvey} which potentially limits industry adaptation.
Models have been found to contradict themselves \cite{2021meoryouPersona} or the grounding, and to add additional information that might be harmful and is not verifiable \cite{languagemodelsthatseekknowledge, 2021hallucinationSurvey}.
Recently, different mitigation strategies were proposed.
\citet{lamda2022} for example use learned discriminators to decide from an n-best list while \citet{rashkin2021ctrl} introduce special control tokens \cite{keskarCTRL2019} to encourage lexical overlap and entailment between grounding and response.
However, discriminators based on estimating human judgments require additional data for training \cite{lamda2022} which is costly to obtain.
Furthermore, discriminating based on single attributes holds the potential of harming other relevant properties.
For example, discriminating based on factuality can lead to responses mostly repeating their grounding information \cite{lamda2022} and might introduce a loss of fluency and dialog coherence.
In general, these two goals might be conflicting and their importance depends on the task at hand.
While chit-chat lives from engagement and might tolerate inconsistencies, factuality is service-critical in task-oriented systems.

In this work, we present a probabilistic model that inherently combines both of these goals.
By factorizing the model according to Bayes' theorem, we obtain one component that models each goal explicitly.
Hence, introducing scaling factors allows for controlling between them. %
Furthermore, additional unlabeled dialog data can be integrated easily to train one of its components.
As directly decoding the model is intractable, we present different approximate decoding schemes for reranking and online decoding that yield significant gains in terms of automatic factuality metrics on several datasets.

\section{Related Work}
\subsection{Document-grounded dialog systems}
There has been significant work in document-grounded dialog systems in recent years.
A large number of datasets have been proposed for open-domain dialog, in order to facilitate engaging conversations about a variety of topics, such as movies \cite{cmu_dog_emnlp18}, Wikipedia knowledge \cite{dinan2018wizard, dziri2022faithdial}, personal attributes of the agent \cite{zhang-etal-2018-personalizing, dinan2020second} or arbitrary information from the internet \cite{2021internetaugmented}.
Similarly, different task-oriented dialog datasets for information-seeking conversations have been proposed \cite{kim-etal-2020-beyond, kim2021howrobustru, feng-2021-dialdoc}.
Different works have dealt with the problem of document retrieval, for example on batching hard negatives \cite{heLearningSelectExternal2021} or efficient document retrieval \cite{thulkeDSTC2021},
as well as with identifying \cite{feng-2021-dialdoc} or rephrasing salient passages within them \cite{languagemodelsthatseekknowledge}.
Finally, there is also significant work on generating grounded responses using this information, for example in low-resource scenarios \cite{zhao2020low} or with an emphasis on faithful generations, which we will explore in the following section.

\subsection{Hallucination in language generation and dialog}
The problem of hallucinations, which one might define as information that is not grounded in the document, dialog context or by common sense, has recently received plenty of attention in neural language generation \cite{2021hallucinationSurvey}, for example in the field of summarization \cite{cao2018faithful, maynez2020faithfulness} and dialog systems \cite{roller2021recipes}.
Hence, different types of mitigation strategies that aim to increase the faithfulness of responses have been proposed.
Notably, \citet{gabriel2021discourse} use a set of discriminators to rerank outputs from an n-best list for summarization.
For dialog systems, \citet{shuster-etal-2021-retrieval-augmentation} show that retrieving relevant information can reduce hallucinations.
However, \citet{romewasbuiltin1776} show that correct grounding does not guarantee faithful outputs.
\citet{lamda2022} train discriminators for dialog systems using human judgements, for example to encourage better grounded responses by adding high-quality responses found by the discriminator to the training data.
\citet{rashkin2021ctrl} augment the input of a grounded response generation model by additional control tokens \cite{keskarCTRL2019} to steer generations towards responses entailed by the grounding and \citet{prabhumoye2021focused} add an additional attention mechanism to BART that focuses solely on the document.

Along with mitigation strategies, methods for model output and metric evaluation for factuality have been proposed.
For example, $Q^2$ \cite{honovich2021q} proposes a question-answering-based matching and BEGIN \cite{dziri2021begin} a benchmark for metric evaluation.

Recently, \citet{dziri2022origin} also show that current grounded datasets contain ground-truth responses that further encourage hallucination by being insufficiently grounded and \citet{dziri2022faithdial} propose FaithDial as a filtered version of Wizard-of-Wikipedia \cite{dinan2018wizard} that aims to mitigate this.

\subsection{Noisy Channel Modeling in NLP}
\label{relatedwork}
Given an input sequence $x_1^T$ and output $y_1^N$, the noisy channel approach \cite{shannonnoisychannel} models the posterior probability of $y_1^N$ given $x_1^T$ as $p\left(y_1^N\mid x_1^T\right) = {p\left(x_1^T \mid y_1^N \right) p\left(y_1^N\right)} / {p\left(x_1^T\right)}$. 
For a long time, such models have been the dominant way of performing Automatic Speech Recognition (ASR) and Machine Translation (MT) \cite{brown-etal-1993-mathematics}.
In ASR, $p(x_1^T \mid y_1^N)$ models the acoustic channel \cite{bahl_jelinek} and is often called channel model.
With the advent of deep learning, discriminative approaches have become popular in both fields and achieve state-of-the-art results \cite{ctc, rnnt, 2017vaswaniAttention, conformer}.
Nevertheless, the noisy channel approach has recently been explored again for MT \cite{yuNeuralNoisyChannel2017, yee2019simple, yuBetterDocumentLevelMachine2020, jeanLogLinearReformulationNoisy2020, subramanian2021nvidia}, text classification \cite{minNoisyChannelLanguage2021}, style transfer \cite{thulke2022dstc} and task-oriented dialog systems that are not document-grounded \cite{liu2021pretraining}. %

\section{Grounded Response Generation}
\label{grounded_generation}
The goal of dialog systems is to find an appropriate system response $u_{T+1}$ conditioned on a sequence of previous turns $u_1^T \coloneqq (u_1, \dots, u_t, \dots, u_T)$ taken by different interlocutors, where each turn $u_t = [u_t]_0^{N_t} \coloneqq ([u_t]_0, \dots [u_t]_{N_t})$ is a sequence of $N_t$ tokens from the model vocabulary $\mathcal{V}$ prepended with the start of sequence symbol $[u_{T+1}]_0 \coloneqq \langle sos \rangle $.
This is usually done by means of a probabilistic language generation model that models the posterior distribution of the response given the context and is locally-normalized such that the response is generated autoregressively according to \begin{align}
    \label{eq:1}
    \begin{split}
        &p\left(u_{T+1} \mid u_1^T \right) \\
        &= \prod_{n=1}^{N_{T+1}} p\left(\left[u_{T+1}\right]_n \mid \left[u_{T+1}\right]_0^{n-1}, u_1^T \right) \text{.}
    \end{split}
\end{align}

In document-grounded dialog systems the response is additionally grounded in information that defines parts of its meaning and is given in the form of unstructured text.
For example, in a restaurant booking setting, text on websites might show whether dogs can be brought or in a chit-chat system the grounding information might define the agent's persona.
While in the latter the grounding is known at test-time as a property of the agent, it is usually unknown in the former case.
Then, the dialog system has to decide for relevant grounding documents, which are obtained from a document base $\mathcal{D}$. %
In this case, retrieval models that model the distribution $p\left(d\mid u_1^T, \mathcal{D}\right)$ in order to rank documents $d \in \mathcal{D}$ may be employed.
By introducing $d$ as a latent variable, the posterior distribution of the response given $u_1^T$ and $\mathcal{D}$ is given as follows.
\begin{align*}
    &p\left(u_{T+1} \mid u_1^T, \mathcal{D} \right) \\ &= \sum_{d \in \mathcal{D}} p \left( u_{T+1}, d \mid u_1^T, \mathcal{D} \right) \\
    &= \sum_{d \in \mathcal{D}} p\left( d \mid u_1^T, \mathcal{D} \right) \cdot p\left( u_{T+1} \mid u_1^T, d, \mathcal{D} \right) \\
    &\approx \sum_{d \in \mathcal{D}} p\left( d \mid u_1^T , \mathcal{D} \right) \cdot p\left( u_{T+1} \mid u_1^T, d\right) %
\end{align*}
For large $\mathcal{D}$ the sum is then approximated by either top-k \cite{rag, thulkeDSTC2021} or maximum approximation.
Furthermore, due to the input length restriction in current language generation models, the dependency on $\mathcal{D}$ is usually dropped, as outlined in the last step.
Using maximum approximation the model becomes
\begin{align*}
    &\hphantom{=} \sum_{d \in \mathcal{D}} p\left( d \mid u_1^T, \mathcal{D} \right) \cdot p\left( u_{T+1} \mid u_1^T, d \right)\\
    &\approx \max_{d \in \mathcal{D}} \left\{ p\left( d \mid u_1^T, \mathcal{D} \right) \cdot p\left( u_{T+1} \mid u_1^T, d \right)\right\} \\
    &\approx p(\hat{d} \mid u_1^T, \mathcal{D}) \cdot p( u_{T+1} \mid u_1^T, \hat{d}) \text{ ,}
\end{align*}
where $\hat{d}$ is the argument of the maximization of just $p( d \mid u_1^T, \mathcal{D} )$ over $\mathcal{D}$ since the simultaneous maximization over both components is intractable.
This results in a widely-used two-step approach (e.g. \cite{kim-etal-2020-beyond, heLearningSelectExternal2021}), where retrieval using $p(d\mid u_1^T, \mathcal{D})$ is followed by a response generation model that uses the retrieved document $\hat{d}$.
Then, since the retrieval probability is constant during response generation, we may decide for a response according to the decision rule
\begin{equation}
     \label{eq:2}
    (u_1^T, d) \mapsto \hat{u}_{T+1} = \argmax_{u_{T+1} \in \mathcal{V}^\ast} \left\{  p(u_{T+1} \mid u_1^T, d) \right\} \text{.}
\end{equation}

In line with recent work \cite{yee2019simple, liu2021pretraining}, we will refer to the model used in \Cref{eq:2} as \emph{direct model} in the following.
While comparatively simple to use, we note that the model can not use data without grounding annotations directly and has been observed to produce a significant number of incorrect outputs, even with ground-truth grounding \cite{romewasbuiltin1776, lamda2022}.

\section{A Noisy Channel Approach}
\label{sec:noisy_channel}
We may use the following equivalent decision rule\footnote{A short proof can be found in \Cref{nc_proof}. Here, equivalence is meant with respect to the true distributions.}, where the direct model is factorized according to Bayes' Theorem:
\begin{align}
    \begin{split}
        &(u_1^T, d) \mapsto \hat{u}_{T+1} = \argmax_{u_{T+1} \in \mathcal{V}^\ast} \left\{  p(u_{T+1} \mid u_1^T, d) \right\} \\
        &= \argmax_{u_{T+1} \in \mathcal{V}^\ast} \big\{ \underbrace{p(d \mid u_{T+1}, u_1^T)}_{\text{"channel model"}} \cdot \underbrace{p(u_{T+1} \mid u_1^T)}_{\substack{\text{"response generation}\\ \text{model"}}} \big\} \text{.}
    \end{split}
\end{align}
Then, the first component can be formulated as a language generation model
\begin{equation}
    \begin{split}
        &p(d_1^M \mid u_{T+1}, u_1^T) \\
    &\hphantom{++}= \prod_{m=1}^{M} p(d_m \mid d_0^{m-1}, u_{T+1}, u_1^T)
    \end{split}
\end{equation}
The first component can be interpreted as favouring responses that allow to reconstruct the grounding based on the generated response.
Hence, one would expect accurate responses to receive higher probability estimates.
The second component is an ungrounded response generation model as in \Cref{eq:1} and favours fluent responses irregardless of the grounding.
Furthermore, it can be trained on large amounts of additional dialogues without grounding annotations. %
In line with previous work on similar models, we will refer to the first component as \emph{channel model}. %
Introducing scaling factors between both components then allows for weighting the two objectives of correctness and fluency in order to control the outputs of the model.

We also note that one might arrive at a similar model by the following computation starting from the document-grounded dialog model without maximum approximation:
\begin{equation}
\label{noisy_channel_alternative}
    \begin{split}
        &p\left(u_{T+1} \mid u_1^T, \mathcal{D} \right) \\ 
        &= \sum_{d \in \mathcal{D}} p\left( d \mid u_1^T, \mathcal{D} \right) \cdot p\left( u_{T+1} \mid u_1^T, d, \mathcal{D} \right) \\
        &= \sum_{d \in \mathcal{D}} \resizebox{0.8\columnwidth}{!}{$p\left( d \mid u_1^T, \mathcal{D} \right) \cdot \frac{p\left(d \mid u_{T+1}, u_1^T, \mathcal{D} \right) \cdot p\left(u_{T+1} \mid u_1^T, \mathcal{D} \right)}{p\left( d \mid u_1^T, \mathcal{D} \right)}$} \\
        &= \sum_{d \in \mathcal{D}} p\left(d \mid u_{T+1}, u_1^T, \mathcal{D} \right) \cdot p\left(u_{T+1} \mid u_1^T, \mathcal{D} \right) \text{.}
    \end{split}
\end{equation}
Nevertheless, we leave the exploration of this model to future work.

\paragraph{Decoding the Noisy Channel Model}

Since the channel model has to be evaluated for each hypothesis generated by the response generation model, decoding the model even with beam search is intractable, as $k \cdot |\mathcal{V}|$-many hypotheses would need to be scored at each iteration with a beam size of $k$.
Therefore, we derive two algorithms to approximately decode the model.
First, we introduce \emph{reranking}, where the noisy channel model is used to score a set of full candidate responses.
Then, we introduce \emph{online decoding}, where the noisy channel model is used to score partial responses during beam search.
In both cases, we resort to a proposal model $q$ to generate candidates.
Recall, that we have the following relationship between direct and noisy channel model:
\begin{equation}
    \resizebox{\columnwidth}{!}{$p(u_{T+1}\mid u_1^T, d) = \frac{p(d\mid u_{T+1}, u_1^T) \cdot p(u_{T+1} \mid u_1^T)}{p(d \mid u_1^T)}$ \text{.}}
\end{equation}
During decoding with the maximum-approximated model introduced in the previous section, $p(d \mid u_1^T)$ is constant and may be dropped.
This makes the direct model the natural choice for a proposal model as we would get the same results as with the noisy channel model given the true distributions.

In reranking, the direct model then generates a set of full responses $U_{T+1}$, from which we decide according to the noisy channel model as follows:
\begin{align}
    \begin{split}
        \hat{u}_{T+1} = \argmax_{u_{T+1} \in U_{T+1}} \Big\{ &p\left(d \mid u_{T+1}, u_1^T\right)^{\lambda_1} \\
        &\cdot p\left(u_{T+1} \mid u_1^T \right)^{\lambda_2} \Big\} \text{,}
    \end{split}
\end{align}
where $\lambda_1 \in \mathbb{R}_{\geq 0}$ and $\lambda_2 \in \mathbb{R}_{\geq 0}$ are scaling factors.
One might also add the direct model without additional computational effort (the probabilities are already calculated during beam search) as an additional factor which has shown beneficial in earlier works \cite{liu2021pretraining}.
We note that this resembles the use of discriminators to select responses but does not require additional annotation effort beyond the grounding annotations.

\begin{algorithm}[t]
\footnotesize
\textbf{Input} Grounding $d$, dialog context $u_1^T$, beam sizes $k_1, k_2$ \newline
\textbf{Output} Response $u_{T+1}$
\begin{algorithmic}[1]
\State Beam: $\mathcal{B} = \{ \langle sos \rangle \}$
\State score$(w) = \log p(w \mid u_1^T, d) \newline
    \hspace*{6em} + {\lambda_1} \cdot \log p(d \mid w, u_1^T) \newline
    \hspace*{8em} +  {\lambda_2} \cdot \log p(w \mid u_1^T) $ \newline 
\State $q(v, w) = \log p(v \mid w, u_1^T, d)$ \newline
\While{\normalfont\texttt{end}($\mathcal{B}$) is \textbf{False}}
    \State \textcolor{brown}{$\mathcal{B}^\prime = \emptyset$}
    \begingroup
    \color{brown}
    \For{$w \in \mathcal{B}$}
        \State \textcolor{brown}{$\mathcal{B}^\prime = \mathcal{B}^\prime \cup \{ w \circ v \mid v \in \displaystyle{\topk_{v \in \mathcal{V}}}  \{ q(v, w\} \}$}\tikzmark{right}
    \EndFor
    \State \textcolor{brown}{$\mathcal{B} = \displaystyle{\topkk_{w \in \mathcal{B}^\prime}} \{ \text{score}(w)\}$}
    \endgroup
    \tikzmark{bottom}
    \State \textcolor{teal}{$\mathcal{B} =  \displaystyle{\topkk_{\substack{w \circ v : \\ w \in \mathcal{B}, v \in \mathcal{V}}}} \{
    q(v, w) + score(w)\}$}
\EndWhile
    \State $\hat{u}_{T+1} = \displaystyle{\argmax_{u_{T+1} \in \mathcal{B}}} \left\{ \frac{\text{score}(u_{T+1})}{\text{len}(u_{T+1})}\right\}$
\end{algorithmic}
\caption{Pseudocode for \textcolor{brown}{Liu et al.'s (\texttt{-})}\\ \cite{liu2021pretraining} and \textcolor{teal}{our (\texttt{+})} decoding algorithm. \label{alg:decoding}}
\end{algorithm}

Since reranking is limited by the hypotheses generated by the proposal model, we propose an online decoding algorithm in which the noisy channel model is used during beam search.
Since the channel model $p(d \mid u_{T+1}, u_1^T)$ depends on the final hypothesis that is not available during search, we approximate it using a model $p(d \mid [u_{T+1}]_0^{n}, u_1^T)$ that only depends on partially generated hypotheses similar to \citet{liu2021pretraining}.
At each step, we score the $k$ hypotheses in our beam using the noisy channel model.
Since scoring all $k\cdot |V|$ possible extensions is infeasible, we select the best $k$ extensions only using their direct model score.
The noisy channel model score for each of the $k$ partial hypotheses $[u_{T+1}]_0^{n}$ up to length $n$ is then calculated as follows:
\begin{align}
    \begin{split}
        p(d \mid [u_{T+1}]_0^{n}, u_1^T)^{\lambda_1} 
        \cdot p([u_{T+1}]_0^{n} \mid u_1^T)^{\lambda_2} \text{.}
    \end{split}
\end{align}
Again, we might add the direct model as an additional factor to the score.
The algorithm requires locally-normalized models and a channel model trained on partial responses, which we obtain by truncating responses according to a uniform distribution over their length in training.

Lastly, we experiment with the algorithm proposed by \citet{liu2021pretraining} which uses the direct model to generate $k_1$ extensions to each of the $k_2$ hypotheses in the beam. The $k_1 \cdot k_2$ new hypotheses are then pruned back to size $k_2$ using the noisy channel model.
We outline both algorithms in Algorithm \ref{alg:decoding}.

\begin{table*}[t]
    \footnotesize
    \resizebox{\textwidth}{!}{\begin{tabular}{l|cc:ccc|cc:ccc}
        \hline
        \multirow{3}{*}{Method} & \multicolumn{5}{c}{ \tabincell{c}{\textbf{Personachat}} } & \multicolumn{5}{c}{ \tabincell{c}{\textbf{FaithDial}}}
        \\
        & sBLEU & METEOR & $Q^2$ & BERTScore & F1 & sBLEU & METEOR & $Q^2$ & BERTScore & F1 \\
         & \multicolumn{2}{c:}{$(u_{T+1},\hat{u}_{T+1})$} & \multicolumn{3}{c|}{$(u_{T+1},d)$} & \multicolumn{2}{c:}{$(u_{T+1},\hat{u}_{T+1})$} & \multicolumn{3}{c}{$(u_{T+1},d)$}\\
        \hline
        \footnotesize
        Direct Model & 5.59 & 17.2 & 0.35 & 0.083 & 14.0 & \textBF{15.16} & \textBF{41.0} & 0.86 & 0.605 & 69.9 \\ 
        + Reranking & 5.46 & 17.3 & 0.48 & 0.162 & 22.9 & 14.36 & 40.1 & 0.89 & 0.654 & 74.4 \\
        + Onl Decoding & \textBF{5.60} & \textBF{19.1} & 0.47 & 0.184 & 25.6 & 13.73 & 39.3 & 0.89 & 0.685 & 77.6 \\
        CTRL & 4.87 & 15.6 & 0.63 & 0.256 & 30.6 & 13.65 & 38.0 & 0.92 & 0.725 & 77.8 \\ 
        + Reranking & 4.14 & 14.4 & \textBF{0.70} & 0.303 & 35.7 & 13.19 & 37.9 & \textBF{0.93} & \textBF{0.749} & 80.6 \\
        + Onl. Decoding & 4.65 & 16.5 & 0.65 & \textBF{0.320} & \textBF{40.1} & 12.70 & 37.1 & 0.92 & \textBF{0.749} & \textBF{81.4} \\
        \multirow{1}{*}{} & \multicolumn{5}{c}{ \tabincell{c}{\textbf{WoW test\_unseen}} } & \multicolumn{5}{c}{ \tabincell{c}{\textbf{WoW test\_seen}}}
        \\
        \hline
        Direct Model & 18.68 & \textBF{37.7} & 0.87 & 0.661 & 74.3 & 18.32 & 36.9 & 0.86 & 0.670 & 74.5 \\ 
        + Reranking & \textBF{18.82} & 37.6 & 0.90 & 0.707 & 78.5 & \textBF{18.47} & 36.8 & 0.89 & 0.702 & 75.3 \\
        + Onl. Decoding & 17.98 & 37.1 & 0.89 & 0.709 & 79.6 & 17.98 & \textBF{37.1} & 0.89 & 0.720 & 79.8 \\
        CTRL & 17.95 & 35.8 & \textBF{0.93} & 0.715 & 77.2 & 17.49 & 35.0 & \textBF{0.92} & 0.729 & 78.0 \\ 
        + Reranking & 17.89 & 35.8 & \textBF{0.93} & 0.740 & 79.8 & 17.67 & 35.0 & \textBF{0.92} & 0.746 & 80.1 \\
        + Onl. Decoding & 17.14 & 35.3 & 0.92 & \textBF{0.743} & \textBF{81.3} & 17.38 & 35.3 & \textBF{0.92} & \textBF{0.766} & \textBF{82.6} \\
        \multirow{1}{*}{} & \multicolumn{5}{c}{ \tabincell{c}{\textbf{DSTC9}} } & \multicolumn{5}{c}{ \tabincell{c}{\textbf{Doc2Dial}}}
        \\ \hline
        Direct Model & 18.60 & 41.5 & 0.71 & 0.439 & 61.9 & 50.08 & \textBF{63.0} & 0.78 & 0.680 & 80.4 \\ 
        + Reranking & 18.33 & 41.3 & 0.76 & 0.489 & 68.4 & 49.90 & 62.3 & 0.83 & 0.725 & 84.3 \\
        + Onl. Decoding & 17.72 & 42.4 & 0.75 & 0.473 & 68.8 & 49.34 & 62.9 & 0.80 & 0.713 & 83.3 \\
        CTRL & \textBF{18.63} & 42.6 & 0.76 & 0.481 & 70.3 & 50.30 & 61.8 & 0.85 & 0.754 & 85.9 \\ 
        + Reranking & 18.45 & 41.6 & \textBF{0.81} & \textBF{0.531} & \textBF{75.2} & \textBF{50.31} & 61.4 & \textBF{0.89} &\textBF{0.796} & \textBF{88.7} \\
        + Onl. Decoding & 17.92 & \textBF{42.8} & 0.79 & 0.506 & 74.3 & 49.11 & 61.4 & 0.87 & 0.764 & 87.5 \\
    
        \hline
    \end{tabular}}
    \caption{Main results of our model compared to the direct model and CTRL \cite{rashkin2021ctrl}. We use our online decoding algorithm and all results are within an effective beam size of 30. \label{main_results}}
\end{table*}

\section{Experiments}
We evaluate our model on multiple different English document-grounded dialog datasets by comparing it to the direct modeling objective and the CTRL model \cite{keskarCTRL2019} presented by \citet{rashkin2021ctrl}, where the input is augmented by a sequence of control tokens $c_1^m$.
Thus, the model becomes $p(u_{T+1} \mid u_1^T, d, c_1^m)$ and we use the noisy channel model $p(d \mid u_{T+1}, u_1^T, c_1^m) \cdot p(u_{T+1} \mid u_1^T)$ in line with Section \ref{sec:noisy_channel}.
We omit the "objective voice" token, since, for example, Personachat specifically targets conversations where the system responds in first person.
The datasets and metrics are described in Section \ref{datasets} and \ref{metrics}, respectively.
In all our experiments, we finetune the BART-large \cite{lewis2019bart} checkpoint that is provided as part of the huggingface transformers \cite{wolf2020transformers} library, which we further use to implement our experiments.
In order to determine the scaling factors for the Noisy Channel model, we do a hyperparameter sweep across $\lambda_i \in \{0.1, 0.2, \dots, 2.0\}$ on the validation sets and choose the parameters that perform best in terms of $Q^2$.
We use $\lambda_1 = 0.6, \lambda_2 = 0.4$ for online decoding and  $\lambda_1 = 0.5, \lambda_2 = 0.2$ for reranking for all experiments after seeing similar trends on all datasets. %
Furthermore, we always use our proposed online decoding algorithm (see Section \ref{online_decoding_by_budget} for a comparison).
The results obtained with these experiments are discussed in Section \ref{results}.

\subsection{Datasets}
\label{datasets}

This section gives a brief overview of the different datasets used in our experiments which capture a variety of settings, for example
 task-oriented and open-domain dialogs grounded
 in documents of varying lengths.
Dataset statistics can be found in \Cref{sec:appendix_dataset_statistics}.

\paragraph{Personachat}
Personachat \cite{zhang-etal-2018-personalizing} is a crowdsourced open-domain dialog dataset, where dialogs are grounded in persona descriptions that consist of five short sentences.
In our experiments, we use the \emph{self} configuration and evaluate on the validation split. %

\paragraph{Wizard-of-Wikipedia}
Wizard-of-Wikipedia (WoW) \cite{dinan2018wizard} is a crowdsourced open-domain dialog dataset, where turns are grounded in Wikipedia articles.
Only the wizard can access the grounding in order to teach the apprentice but also choose not to use any grounding.
We evaluate on the subset of grounded wizard turns.

\paragraph{DSTC9}
DSTC9 Track 1
is an extension of the MultiWoz 2.1 dataset \cite{eric2019multiwoz}, where turns require information beyond the existing API structure and which was collected from FAQ documents.
The test set contains conversations about a new location and a held-out domain, as well as transcripts of spoken conversations.%

\paragraph{Doc2Dial}
Doc2dial \cite{feng-etal-2020-doc2dial} is a task-oriented dialog dataset, where the agent provides a user with information from public government service websites.
The grounding annotations are given on different levels.
We use a concatenation of the annotated spans as a grounding instead of taking entire paragraphs.

\paragraph{FaithDial}
Based on the observation that current dialog datasets contain insufficiently grounded annotations that encourage hallucinations \cite{dziri2022origin}, \citet{dziri2022faithdial} release FaithDial, an edited version of Wizard-of-Wikipedia that contains significantly less hallucinations.
For this, crowdworkers have edited 44\% of the grounded training responses from seeker-initiated conversations and all those from the validation and test set.

\subsection{Evaluation Metrics}
\label{metrics}
In line with the shared tasks on some of the datasets we use for evaluation \cite{kim-etal-2020-beyond, kim2021howrobustru, feng-etal-2020-doc2dial}, we use the sacrebleu (sBLEU) \cite{sacrebleu} implementation of BLEU \cite{papineni2002bleu}, and METEOR \cite{banerjee2005meteor} to assess our model generations with word-overlap based metrics.
In addition to that, we use BERTScore \cite{bert-score} and the token-level F1-Score between $u_{T+1}$ and $d$, as well as the recently proposed $Q^2$ metric \cite{honovich2021q} to evaluate the factual consistency of our models.
$Q^2$ is a model-based metric that matches the answers, which are derived from response and grounding, to questions generated based on the response using an NLI model and has shown strong correlations with human judgements in system-level evaluation on WoW.

\subsection{Retrieval}
In addition to experiments that use ground-truth grounding, we also experiment with using the outputs of retrieval models, since usually the grounding is not known at inference time.
For retrieval we use two architectures.
First, a Bi-Encoder \cite{bromley1993signature}, where a dialog and document encoder model map $u_1^T$ and each $d \in \mathcal{D}$ to a fixed-size dense vector of the same dimension, respectively.
The grounding document is determined by nearest neighbor search, i.e. the decision is made for the document whose vector is closest to the dialog vector in the embedding space.
In our experiments, the weights of dialog and document encoder are shared and trained using the Triplet loss criterion.

Secondly, we employ a Cross-Encoder which provides strong performance across a variety of tasks but remains too inefficient in order to be used in practice with large $\mathcal{D}$ \cite{reimers2019sentence, humeau2019poly, karpukhin2020dense, thulkeDSTC2021}.
In the Cross-Encoder, dialog context and document are concatenated as the input to a Transformer model that subsequently performs relevance classification such that the document with the highest score is retrieved.

We use RoBERTa-large \cite{liu2019roberta} for all experiments use Recall@1 (R@1) for evaluation.

\section{Results}
\label{results}
\begin{figure*}[t]
\centering
\begin{minipage}{.33\textwidth}
  \resizebox{\columnwidth}{!}{\begin{tikzpicture}
    \begin{axis}[
      xmin=0,xmax=45,
      ymin=69,ymax=99,
      axis y line*=left,
      xlabel={Effective beam size},
      xlabel near ticks,
      ylabel={$Q^2 \cdot 100$},
      ylabel near ticks
    ]
    \addplot+[black,mark=star,mark options={fill=gray}, domain=35:0] coordinates {(50, 100)};
  \label{pgfplots:plot1}
    \addplot+[teal,mark=o,mark options={fill=gray}, domain=35:0] coordinates {
    (40, 88.2)
    (35, 87.8)
    (30, 87.5)
    (25, 87.5)
    (20, 87.4)
    (16, 86.2)
    (12, 85.8)
    (9, 83.8)
    (6, 81.6)
  };
    \addplot+[brown,mark=asterisk,mark options={fill=gray}, domain=35:0] coordinates {
    (40, 82.9)
    (35, 81.8)
    (30, 81.5)
    (25, 80.7)
    (20, 81.0)
    (16, 79.6)
    (12, 80.2)
    (9, 80.3)
    (6, 80.3)
  };
      \addplot+[lightgray,mark=|,mark options={fill=gray}, domain=35:0] coordinates {
    (40, 85.2)
    (35, 84.9)
    (30, 85.3)
    (25, 85.0)
    (20, 85.9)
    (16, 84.1)
    (12, 83.1)
    (9, 83.0)
    (6, 80.5)
  };
    \end{axis}
    \begin{axis}[
      xmin = 0, xmax = 45,
      ymin = 69, ymax = 69,
      hide x axis,
      axis y line*=right,
      yticklabels={,,}
    ]
   \node[draw=black,fill=white,anchor=north west] at (rel axis cs:0,1) {WoW test\_seen};
    \end{axis}
  \end{tikzpicture}}
\end{minipage}%
\begin{minipage}{.33\textwidth}
  \centering
  \resizebox{\columnwidth}{!}{\begin{tikzpicture}
    \begin{axis}[
      xmin=0,xmax=45,
      ymin=29,ymax=59,
      axis y line*=left,
      xlabel={Effective beam size},
      xlabel near ticks,
      ylabel={$Q^2 \cdot 100$},
      ylabel near ticks
    ]
   \node[draw=black,fill=white,anchor=north west] at (rel axis cs:0,1) {Personachat};
    \addplot+[black,mark=o,mark options={fill=gray}, domain=35:0] coordinates {(50, 100)};
  \label{pgfplots:plot1}
    \addplot+[teal,mark=o,mark options={fill=gray}, domain=35:0] coordinates {
    (40, 47.6)
    (35, 46.8)
    (30, 45.9)
    (25, 45.1)
    (20, 44.1)
    (16, 43.4)
    (12, 42.4)
    (9, 41.2)
    (6, 40.4)
  };
    \addplot+[brown,mark=asterisk,mark options={fill=gray}, domain=35:0] coordinates {
    (40, 40.9)
    (35, 40.3)
    (30, 39.7)
    (25, 39.8)
    (20, 39.8)
    (16, 39.5)
    (12, 39.6)
    (9, 39.3)
    (6, 39.9)
  };
      \addplot+[lightgray,mark=|,mark options={fill=gray}, domain=35:0] coordinates {
    (40, 35.6)
    (35, 35.7)
    (30, 35.5)
    (25, 35.4)
    (20, 35.5)
    (16, 35.4)
    (12, 35.6)
    (9, 35.5)
    (6, 36.2)
  };
    \end{axis}
    \begin{axis}[
      xmin = 0, xmax = 45,
      ymin = 29, ymax = 59,
      hide x axis,
      axis y line*=right,
      yticklabels={,,}
    ]
    \addplot+[black,mark=triangle*,mark options={fill=gray}, domain=35:0] coordinates {(50, 100)};
    \end{axis}
  \end{tikzpicture}}
\end{minipage}
\begin{minipage}{.33\textwidth}
  \centering
  \resizebox{\columnwidth}{!}{\begin{tikzpicture}
    \begin{axis}[
      xmin=0,xmax=45,
      ymin=59,ymax=89,
      axis y line*=left,
      xlabel={Effective beam size},
      xlabel near ticks,
      ylabel={$Q^2 \cdot 100$},
      ylabel near ticks
    ]
   \node[draw=black,fill=white,anchor=north west] at (rel axis cs:0,1) {DSTC9};
    \addplot+[black,mark=*,mark options={fill=gray}, domain=35:0] coordinates {(50, 100)};
  \label{pgfplots:plot1}
    \addplot+[teal,mark=o,mark options={fill=gray}, domain=35:0] coordinates {
    (40, 76.4)
    (35, 76.3)
    (30, 76.5)
    (25, 76.4)
    (20, 76.5)
    (16, 76.1)
    (12, 75.6)
    (9, 75.5)
    (6, 74.6)
  };
    \addplot+[brown,mark=asterisk,mark options={fill=gray}, domain=35:0] coordinates {
    (40, 71.3)
    (35, 71.2)
    (30, 71.1)
    (25, 71.1)
    (20, 70.9)
    (16, 70.6)
    (12, 70.6)
    (9, 70.0)
    (6, 70.3)
  };
      \addplot+[lightgray,mark=|,mark options={fill=gray}, domain=35:0] coordinates {
    (40, 69.4)
    (35, 69.5)
    (30, 69.4)
    (25, 69.8)
    (20, 70.2)
    (16, 70.6)
    (12, 70.8)
    (9, 71.3)
    (6, 71.8)
  };
    \end{axis}
    \begin{axis}[
      xmin = 0, xmax = 45,
      ymin = 59, ymax = 89,
      hide x axis,
      axis y line*=right,
    yticklabels={,,}
    ]
    \addplot+[black,mark=triangle*,mark options={fill=gray}, domain=35:0] coordinates {(50, 100)};
    \end{axis}
  \end{tikzpicture}}
\end{minipage}
\end{figure*}
\begin{figure*}[t]
\centering
\begin{minipage}{.33\textwidth}
  \resizebox{\columnwidth}{!}{\begin{tikzpicture}
    \begin{axis}[
      xmin=0,xmax=45,
      ymin=0,ymax=29,
      axis y line*=left,
      xlabel={Effective beam size},
      xlabel near ticks,
      ylabel={sBLEU\vphantom{$Q^2$}},
      ylabel near ticks
    ]
    \addplot+[black,mark=*,mark options={fill=gray}, domain=35:0] coordinates {(50, 100)};
  \label{pgfplots:plot1}
    \addplot+[teal,mark=o,mark options={fill=gray}, domain=35:0] coordinates {
    (40, 88.2)
    (35, 87.8)
    (30, 87.5)
    (25, 87.5)
    (20, 87.4)
    (16, 86.2)
    (12, 85.8)
    (9, 83.8)
    (6, 81.6)
  };
    \addplot+[brown,mark=asterisk,mark options={fill=gray}, domain=35:0] coordinates {
    (40, 82.9)
    (35, 81.8)
    (30, 81.5)
    (25, 80.7)
    (20, 81.0)
    (16, 79.6)
    (12, 80.2)
    (9, 80.3)
    (6, 80.3)
  };
      \addplot+[lightgray,mark=|,mark options={fill=gray}, domain=35:0] coordinates {
    (40, 85.2)
    (35, 84.9)
    (30, 85.3)
    (25, 85.0)
    (20, 85.9)
    (16, 84.1)
    (12, 83.1)
    (9, 83.0)
    (6, 80.5)
  };
    \end{axis}
    \begin{axis}[
      xmin = 0, xmax = 45,
      ymin = 0, ymax = 29,
      hide x axis,
      axis y line*=right,
      yticklabels={,,}
    ]
   \node[draw=black,fill=white,anchor=north west] at (rel axis cs:0,1) {WoW test\_seen};
    \addplot+[black,mark=triangle*,mark options={fill=gray}, domain=35:0] coordinates {(50, 100)};
      \addplot+[teal,mark=o,mark options={fill=gray}, domain=35:0] coordinates {
      (40, 19.31)
      (35, 19.12)
      (30, 19.04)
      (25, 18.93)
      (20, 18.86)
      (16, 18.69)
      (12, 18.50)
      (9, 18.23)
      (6, 18.17)
      };
    \addplot+[brown,mark=asterisk,mark options={fill=gray}, domain=35:0] coordinates {
      (40, 18.99)
      (35, 18.86)
      (30, 18.86)
      (25, 18.68)
      (20, 18.73)
      (16, 18.64)
      (12, 18.72)
      (9, 18.80)
      (6, 18.95)
     };
      \addplot+[lightgray,mark=|,mark options={fill=gray}, domain=35:0] coordinates {
    (40, 19.04)
    (35, 19.03)
    (30, 18.89)
    (25, 18.79)
    (20, 18.84)
    (16, 18.95)
    (12, 18.69)
    (9, 18.84)
    (6, 18.62)
  };
    \end{axis}
  \end{tikzpicture}}
\end{minipage}%
\begin{minipage}{.33\textwidth}
  \centering
  \resizebox{\columnwidth}{!}{\begin{tikzpicture}
    \begin{axis}[
      xmin=0,xmax=45,
      ymin=5,ymax=10,
      axis y line*=left,
      xlabel={Effective beam size},
      xlabel near ticks,
      ylabel={sBLEU\vphantom{$Q^2$}},
      ylabel near ticks
    ]
   \node[draw=black,fill=white,anchor=north west] at (rel axis cs:0,1) {Personachat};
    \addplot+[black,mark=*,mark options={fill=gray}, domain=35:0] coordinates {(50, 100)};
  \label{pgfplots:plot1}
    \end{axis}
    \begin{axis}[
      xmin = 0, xmax = 45,
      ymin = 0, ymax = 10,
      hide x axis,
      axis y line*=right,
      yticklabels={,,}
    ]
    \addplot+[black,mark=o,mark options={fill=gray}, domain=35:0] coordinates {(50, 100)};
      \addplot+[teal,mark=o,mark options={fill=gray}, domain=35:0] coordinates {
      (40, 5.34)
      (35, 5.45)
      (30, 5.48)
      (25, 5.60)
      (20, 5.64)
      (16, 5.60)
      (12, 5.60)
      (9, 5.50)
      (6, 5.74)
      };
    \addplot+[brown,mark=asterisk,mark options={fill=gray}, domain=35:0] coordinates {
      (40, 5.76)
      (35, 5.69)
      (30, 5.72)
      (25, 5.80)
      (20, 5.80)
      (16, 5.66)
      (12, 5.66)
      (9, 5.55)
      (6, 5.63)
     };
      \addplot+[lightgray,mark=|,mark options={fill=gray}, domain=35:0] coordinates {
    (40, 5.64)
    (35, 5.65)
    (30, 5.57)
    (25, 5.55)
    (20, 5.63)
    (16, 5.66)
    (12, 5.54)
    (9, 5.69)
    (6, 5.67)
  };
    \end{axis}
  \end{tikzpicture}}
\end{minipage}
\begin{minipage}{.33\textwidth}
  \centering
  \resizebox{\columnwidth}{!}{\begin{tikzpicture}
    \begin{axis}[
      xmin=0,xmax=45,
      ymin=0,ymax=29,
      axis y line*=left,
      xlabel={Effective beam size},
      xlabel near ticks,
      ylabel={sBLEU\vphantom{$Q^2$}},
      ylabel near ticks
    ]
   \node[draw=black,fill=white,anchor=north west] at (rel axis cs:0,1) {DSTC9};
    \end{axis}
    \begin{axis}[
      xmin = 0, xmax = 45,
      ymin = 0, ymax = 29,
      hide x axis,
      axis y line*=right,
    yticklabels={,,}
    ]
      \addplot+[teal,mark=o,mark options={fill=gray}, domain=35:0] coordinates {
      (40, 18.37)
      (35, 18.31)
      (30, 18.38)
      (25, 18.40)
      (20, 18.42)
      (16, 18.38)
      (12, 18.37)
      (9, 18.49)
      (6, 18.58)
      };
    \addplot+[brown,mark=asterisk,mark options={fill=gray}, domain=35:0] coordinates {
      (40, 18.40)
      (35, 18.49)
      (30, 18.46)
      (25, 18.76)
      (20, 18.79)
      (16, 18.85)
      (12, 18.83)
      (9, 18.65)
      (6, 18.57)
     };
      \addplot+[lightgray,mark=|,mark options={fill=gray}, domain=35:0] coordinates {
    (40, 17.82)
    (35, 17.84)
    (30, 17.83)
    (25, 17.86)
    (20, 17.98)
    (16, 18.07)
    (12, 18.42)
    (9, 18.61)
    (6, 18.94)
  };
    \end{axis}
  \end{tikzpicture}}
\end{minipage}
    \caption{$Q^2$ and sBLEU for the noisy channel model using \colorbox{teal!30}{our decoding (o)}, the one proposed by \colorbox{brown!30}{Liu et al. 2019 (*)}, where $k_1\cdot k_2$ is the effective beam size, and \colorbox{lightgray!30}{direct model beam search (|)} by compute budget. \label{tab:decoding}}
\end{figure*}
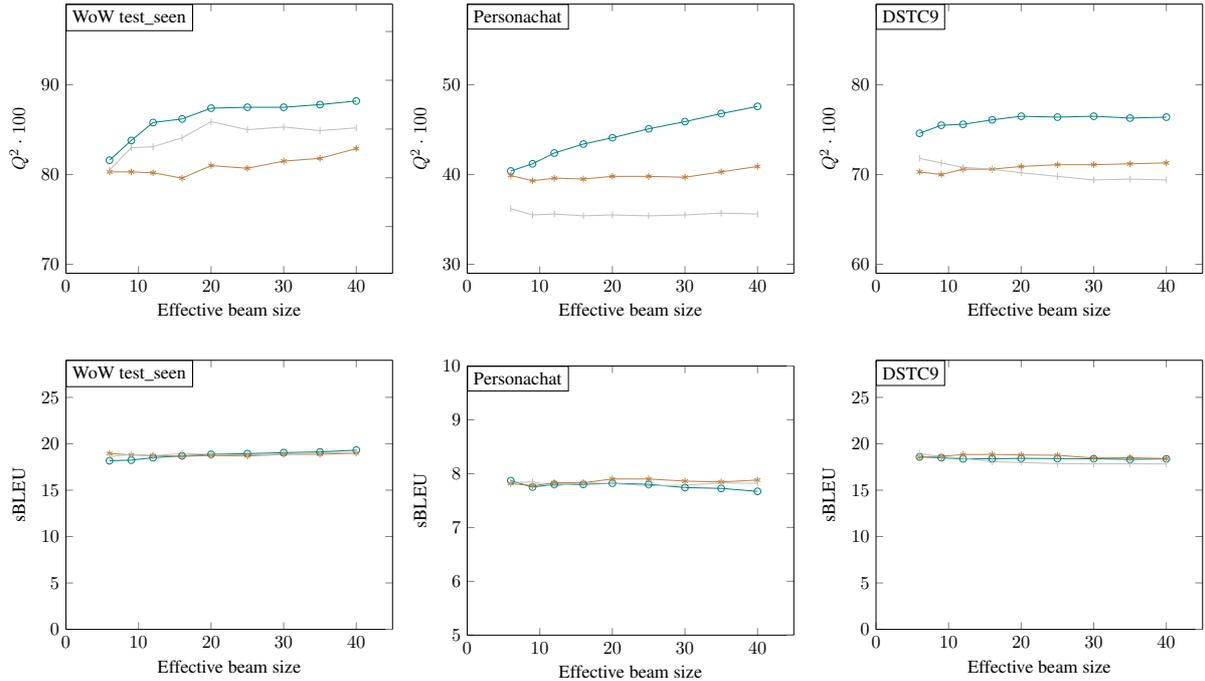

Table \ref{main_results} shows the results obtained with our model using our proposed online decoding algorithm, reranking, and no additional data in training but the corresponding training set.
We identify the following trends:

1) Our model consistently outperforms the direct modeling objective in terms of all automated factuality metrics.

2) There is no clear trend in terms of word-overlap-based metrics, where our model and the direct model show comparable performance.

3) CTRL gives larger improvements in terms of $Q^2$ on all datasets but DSTC9. Nevertheless, the additional control tokens may be seen as a data filtering method that adds new information to the training data that is not available to our model.

4) Adding control tokens to our model, i.e. combining CTRL with the noisy channel approach, gives further improvements in terms of factuality metrics.

5) The improvements obtained on the unseen set of Wizard-of-Wikipedia indicate that the model is also able to generalize appropriately to new information.

 Overall, the results indicate that by scaling the channel model contribution appropriately, the faithfulness of responses can indeed be improved in comparison to the direct model.
In the following, we present further results to understand how the model behaves under different scaling factors (Section \ref{section:controllability}), uncertain retrieval (Section \ref{section:retrieval}), different compute budgets (Section \ref{online_decoding_by_budget}), and the presence of additional data (Section \ref{section:additional_data}) before concluding the section with a qualitative analysis (Section \ref{sec:qualitative}).
\subsection{Controllability}
\label{section:controllability}
\begin{table}
\small
    \centering
    \resizebox{\columnwidth}{!}{\begin{tabular}{l|cc:ccc}
        \hline 
         $\lambda_1 / \lambda_2$ & sBLEU $\uparrow$ & PPL$\downarrow$ & $Q^2\uparrow$  & BERTScore $\uparrow$ & LCS  \\
         & \multicolumn{2}{c:}{$(u_{T+1},\hat{u}_{T+1})$} & & ($u_{T+1}, d$) &  \\ \hline
        Direct Model & 5.59 & 9.6 & 0.36 & 0.08 & 0.33 \\ \hdashline
        0.0 / 0.5 & 5.26 & 9.1 & 0.34 & 0.07 & 0.33 \\
        0.2 / 0.5 & 5.67 & 10.2 & 0.38 & 0.12 & 0.35 \\
        0.4 / 0.5 & 5.52 & 10.7 & 0.40 & 0.13 & 0.36 \\
        0.6 / 0.5 & 5.48 & 11.2 & 0.41 & 0.14 & 0.37 \\
        0.8 / 0.5 & 5.29 & 11.5 & 0.42 & 0.15 & 0.38 \\
        1.0 / 0.5 & 5.07 & 11.9 & 0.43 & 0.16 & 0.38 \\ \hdashline
        1.0 / 5.0 & 3.94 & 8.70 & 0.38 & 0.17 & 0.33 \\
        \hline
        CTRL & 4.89 & 17.2 & 0.64 & 0.27 & 0.66 \\ \hdashline
        0.6 / 0.4 & 4.70 & 19.5 & 0.64 & 0.32 & 0.62 \\
        \hline
        Gold response & 100.0 & 18.3 & 0.26 & 0.09 & 0.25 \\
        Gold document & 1.97 & 47.8 & 99.7 & 1.0 & 1.0 \\
         \hline
    \end{tabular}}
    \caption{$Q^2$ and Perplexity by ratio of factors on Personachat with beam size 10.}
    \label{controllability}
\end{table}
\Cref{controllability} shows results obtained with noisy channel online decoding with different scaling factors. In addition to the previously mentioned metrics, we report Perplexity as a proxy for fluency \cite{dinan2020second} and the longest common subsequence (LCS) of $u_{T+1}$ and $d$ normalized by the length of $u_{T+1}$ to indicate how much of the response is copied from the grounding.
We can see that with a higher channel model factor, the model copies more from the grounding and produces more factual responses.
However, these responses have a higher perplexity, as judged by a response generation model trained on all datasets.
Thus, by differing the scaling factors one might control how much from the grounding is copied and how much importance is given to retaining a response that fits well into the dialog.
CTRL copies significantly more from the grounding than our model.
All models, including the direct model, copy more than the ground-truth responses, which is also consistent across all datasets.
We note that this may not be desirable in open-domain dialog settings.

\subsection{Retrieval}
\label{section:retrieval}
\begin{table}
    \centering
    \resizebox{\columnwidth}{!}{\begin{tabular}{|l|c|cc:ccc|}
    \hline
    Method     & R@1 & sBLEU & \scriptsize{METEOR} & $Q^2$ & \scriptsize{BERTScore} & F1 \\ \hline
    Bi-Encoder & 83.0 & - & - & - & - & - \\ \hdashline
    Direct Model & 83.0 & \textBF{17.87} & 39.8 & 0.66 & 0.403 & 56.2 \\
    Online Decoding & 83.0 & 17.80 & \textBF{41.0} & \textBF{0.68} & \textBF{0.434} & \textBF{61.3} \\ \hline
    Cross-Encoder & 87.2  & - & - & - & - & - \\ \hdashline
    Direct Model & 87.2 & \textBF{18.04} & 40.4 & 0.66 & 0.412 & 57.7 \\
    Online Decoding & 87.2 & 17.83 & \textBF{41.6} & \textBF{0.69} & \textBF{0.446} & \textBF{63.1} \\
   \hline
    \end{tabular}}
    \caption{Results on the outputs of a Bi-Encoder and Cross-Encoder retrieval model on DSTC9 test.
    \label{tab:retrieval}}
\end{table}
Table \ref{tab:retrieval} show results obtained with the noisy channel model and the direct model with uncertain retrieval.
We limit our study to DSTC9, since setting-up competitive retrievers for each dataset is non-trivial. For example, on doc2dial a QA-based pipeline is often used \cite{daheim2021cascaded, feng-2021-dialdoc}).
For both the Bi-Encoder and the better performing Cross-Encoder retrieval model, the noisy channel model produces more factual responses than the direct model.
This shows that the model remains effective in a realistic setting.

\subsection{Online Decoding by compute budget}
\label{online_decoding_by_budget}

Figure \ref{tab:decoding} shows a comparison of our proposed decoding algorithm, the algorithm outlined in \cite{liu2021pretraining} and standard beam search using the direct model. We can see that the noisy channel model with our decoding algorithm consistently gives the best performance in terms of $Q^2$.
Furthermore, the direct model even performs better than the decoding algorithm proposed in \cite{liu2021pretraining} on Wizard-of-Wikipedia for the same compute budget, which takes a significant number of beams for improvements to show.
In general, a larger beam size improves the performance of the noisy channel model but not the direct model. %

\subsection{Additional data}
\label{section:additional_data}
\begin{table}[]
    \centering
    \resizebox{\columnwidth}{!}{\begin{tabular}{|l|cc|cc|cc|}
    \hline
    & \multicolumn{2}{c|}{\textBF{Personachat}} & \multicolumn{2}{c|}{\textBF{DSTC9}} & \multicolumn{2}{c|}{\textBF{FaithDial}} \\ \hline
        Method & sBLEU & $Q^2$ & sBLEU & $Q^2$ & sBLEU & $Q^2$  \\ \hline
        Reranking & 5.46 & \textBF{0.48} & 18.33 & 0.76 & 14.36 & 0.89\\
        + data & \textBF{5.89} & 0.46 & 18.47 & 0.73 & \textBF{15.29} & 0.87\\
        Onl. Decoding & 5.60 & 0.47 & 17.72 & 0.75 & 13.73 & 0.89\\
        + data & 5.60 & 0.46 & \textBF{18.67} & \textBF{0.77} & 13.51 & \textBF{0.91} \\ \hline
    \end{tabular}}
    \caption{Results with noisy channel model and additional response generation model training data.
    \label{tab:additional_data}}
\end{table}
In order to study the effect of additional training data for the response generation model, we train the component on all target datasets as well as MultiWoZ 2.1 \cite{eric2019multiwoz} \footnote{We remove the dialogs contained in DSTC9.}, Taskmaster-1 \cite{byrne-etal-2019-taskmaster}, TopicalChat \cite{Gopalakrishnan2019} and CMU DoG \cite{cmu_dog_emnlp18}.
However, as shown in Table \ref{tab:additional_data}, we do not see consistent but often dataset-specific improvements. For example, $Q^2$ tends to be better with more data in reranking but not online decoding, which also did not change in our experiments with different scaling factors.

\subsection{Qualitative Analysis}
\label{sec:qualitative}
When comparing the generation outputs of all models, we can make the following observations:

1) the outputs of both the noisy channel model and CTRL appear more faithful to the grounding and more specific.
For example, on DSTC9 and doc2dial the direct model sometimes leaves out important details, such as that an ID card needs to be shown when someone picks up a ticket at a train station, which both noisy channel model and CTRL incorporate into the response.
Furthermore, the direct model appears to generate generic responses more often, such as "do you have a pet?" when the topic in Personachat is "dog" or "cat".

2) In general, qualitative analysis supports the results from our automatic evaluation that a higher channel model factor implies more from the grounding being copied into the response, which however can come at a loss of fluency and coherence.

3) A higher response generation model factor leads to more abstractiveness and a better connected response that, for example. contains follow-up questions more often.
On the other hand, a too high factor also led to hallucinations in our experiments.

4) When comparing CTRL and the noisy channel model, the main difference appears to be that the responses of the noisy channel model (with a suitable factor) seem more connected to the previous turns than in CTRL, especially on a chit-chat task like Personachat.
Nevertheless, with a higher channel model factor the generations become more similar to those of CTRL.
The combination of CTRL and the noisy channel model might be especially suited for task-oriented dialog, where faithfulness is crucial, whereas for some open-domain settings we think that our model may be more suitable without control tokens.

Finally, some example outputs can be found in the Appendix.
    
\section{Conclusion}
In this paper, we present a model for response generation in document-grounded dialog that explicitly optimizes for faithfulness and fluency.
The model decomposes the posterior distribution of response given context and grounding into two components according to Bayes' Theorem and thus, by introducing scaling factors, allows for encouraging more correct or more fluent responses.
Since decoding the model directly is intractable, we derive and compare different approximate decoding schemes that use reranking or online decoding.
We compare the model to directly modeling the posterior distribution of response given context and grounding and a variant of CTRL that was proposed recently to encourage faithfulness in grounded response generation, which we again factorize into two components.
An evaluation on five different open-domain and task-oriented dialog datasets shows improvements in terms of factuality on top of both models.
Furthermore, we highlight how the scaling factors can be used to control how much the model copies from the grounding and how much weight is given to a well-connected response, and we investigate the influence of additional training data for one of the model components.
Lastly, we show that our model also gives improvements when dealing with uncertain document retrieval.

\subsection{Future Work}
In future work we would like to explore the noisy channel model for document-grounded response generation without maximum approximation, such as the one that we present in Equation (\ref{noisy_channel_alternative}).

\section{Limitations}
The main limitation of our model is that the approximate decoding schemes introduce significant computational overhead in comparison to the direct model and CTRL, which amounts to a factor of up to 10 for online decoding with a large beam size using a not yet fully optimized implementation. %
Furthermore, decoding is a lot more complex and the scaling factors need to be tuned which implies significant additional computations that are necessary and results in a larger carbon footprint.
In addition, our noisy channel model has a significantly higher number of parameters than the baseline model.
Finally, we mainly rely on automatic metrics to assess the faithfulness of the proposed approaches and leave out a broader evaluation of the general quality of generated responses but for a small-scale qualitative study cf. Section \ref{sec:qualitative}.

\section{Broader Impact}
In general, generative dialog systems are a promising field of research and can be less restricted in the topics they can deal with than rule-based models or approaches that use predefined dialog flows, for example, which require a lot of handcrafting and possibly experts to write appropriate responses.
Nevertheless, such language generation-based approaches bear the danger of repeating harmful content and biases that may have been present in the training data or of generating inappropriate responses, in general.
Furthermore, some applications require faithful responses by law or are at least critical to the service.
While our model can improve the faithfulness of responses, it can not be guaranteed.
Therefore, we would not recommend to use the model in these applications.
While document-grounded dialog systems in general can be used to ground system responses in helpful and correct real-world information, these systems could potentially also be misused to ground dialog systems in misinformation or other harmful documents.

\section*{Acknowledgements}
This work has been financially supported by multiple sources:
The project HYKIST funded by the German Federal Ministry of Health on the basis of a decision of the German Federal Parliament (Bundestag) under funding ID ZMVI1-2520DAT04A.

The project NeuroSys which, as part of the initiative \enquote{Clusters4Future}, is funded by the Federal Ministry of Education and Research BMBF (03ZU1106DA).

The German Federal Ministry of Education and Research and the Hessian Ministry of Higher Education, Research, Science and the Arts within their joint support of the National Research Center for Applied Cybersecurity ATHENE. 

The European Union under the Horizon Europe grant No. 101070351 (SERMAS).

\bibliography{acl_latex}
\bibliographystyle{acl_natbib}

\clearpage
\newpage

\appendix

\section{Appendix}

\subsection{Proof of Equivalence of direct and noisy channel model decoding}
\label{nc_proof}
Given a dialog context $u_1^T$ and grounding $d$, we may define the search problem as finding an appropriate response $u_{T+1}$.
Following our derivation from Section \ref{grounded_generation} we can formalize this according to the decision rule \begin{equation}
    (u_1^T, d) \mapsto \hat{u}_{T+1} = \argmax_{u_{T+1} \in \mathcal{V}^\ast} \{  p(u_{T+1} \mid u_1^T, d) \} \text{,}
\end{equation}
where $\mathcal{V}^\ast$ denotes the set of finite strings that may be constructed from a fixed vocabulary $\mathcal{V}$ using the Kleene closure $\ast$.
Then, we may use the following equivalent decision rule \begin{equation}
\begin{split}
        (u_1^T, d) \mapsto  \hat{u}_{T+1} = &\argmax_{u_{T+1} \in \mathcal{V}^\ast} \{ p(d \mid u_{T+1}, u_1^T) \\
        &\hphantom{argmax} \cdot p(u_{T+1} \mid u_1^T) \} \text{.}
\end{split}
\end{equation}
\begin{proof}
\begin{align*}
    &\argmax_{u_{T+1} \in \mathcal{V}^\ast} \{  p(u_{T+1} \mid u_1^T, d) \} \\
    &=\argmax_{u_{T+1} \in \mathcal{V}^\ast} \{  p(u_{T+1} \mid u_1^T, d) \cdot p(d \mid u_1^T) \} \\
    &=\argmax_{u_{T+1} \in \mathcal{V}^\ast} \{  p(u_{T+1}, d \mid u_1^T) \} \\
    &=\argmax_{u_{T+1} \in \mathcal{V}^\ast} \{  p(d \mid u_{T+1},  u_1^T) \cdot p(u_{T+1} \mid u_1^T) \}\\
    \end{align*}
Here, the first step is obtained by multiplying with $p(d\mid u_1^T)$, which does not change the argument of the maximization.
\end{proof}

\subsection{Experiment details}
For all of our experiments and models, we use BART-large \cite{lewis2019bart}, which adheres to the standard Transformer model from \citet{2017vaswaniAttention} and consists of 12 encoder and 12 decoder layers with a hidden size of 1024 and 406M parameters.
We finetune each model for 10 epochs using an initial learning rate of $6.25e-5$, with no warmup steps, and linear learning rate decay.
We use a batch size of 32 by using gradient accumulation.
The model is evaluated on the validation set after each epoch and the model with the smallest eval loss is picked as our final model.
We truncate the dialog history at 384 tokens and restrict the length of the grounding to 128 tokens, after which it is cut off.

Except for the Bi-Encoder, which we train using the Triplet Loss, all models are trained using the Cross-Entropy criterion.

All of our experiments using online decoding use $\lambda_1 = 0.6, \lambda_2 = 0.4$ and all of the reranking experiments use $\lambda_1 = 0.5, \lambda_2 = 0.2$. 

All models were trained and evaluated on NVIDIA 1080 or 2080 GPU's.

\subsection{Dataset statistics}
\label{sec:appendix_dataset_statistics}
    \begin{table}[H]
        \resizebox{\columnwidth}{!}{\begin{tabular}{|l|rrrr|}
            \hline
            Dataset & Split & Domain & \# Dialogs & \#Documents \\ \hline
            DSTC9& train & Task-oriented & 19,184 & 2,900 \\
             \cite{kim-etal-2020-beyond} & test & & 1,981 & 12,039\\
            Personachat  & train & Open-domain & 10,907 & 1,155 \\
            \cite{zhang-etal-2018-personalizing}& test & & 1,000 &  \\ \hline
            Wizard-of-Wikipedia & train & Open-domain & 18,430 & 93M sentences \\
            \cite{dinan2018wizard} & test\_seen & & 965 & 93M sentences \\
            & test\_unseen & & 968 & 93M sentences \\ \hline
            FaithDial  & train & Open-domain & 4,094 & 93M sentences \\
            \cite{dziri2022faithdial}& test & & 791 & 93M sentences \\ \hline
            Doc2Dial  & train & Task-oriented & 3,474 & 487 \\
            \cite{feng-etal-2020-doc2dial}& test & & 787 & 572 \\ \hline
        \end{tabular}} 
    \end{table}

\subsection{Sample model outputs}
In this section we present a number of cherry-picked examples for multiple datasets to supplement the qualitative analysis from Section \ref{sec:qualitative}.

Furthermore, we present examples obtained with different scaling factors to highlight the behavior of the model under these conditions.

\begin{table*}
\small
 \resizebox{\textwidth}{!}{\begin{tabular}{|l|l|}
    \multicolumn{2}{c}{\textbf{Personachat}} \\
    \hline
    Grounding  & \linebreakcell{i enjoy listening to classical music . i'm a christian . \\ my favorite color is red . i can drive a tractor . my sister is a pole dancer .}
\\
    $u_T$  & yes it is . i am not a fan of high maintenance people as well \\
    Ground truth & i am far form that , i drive a tractor and do not mind taking it easy \\ \hline
    Direct Model & what do you do for a living? \\ \hdashline
    Reranking & do you have any hobbies? i like to drive a tractor \\
    Noisy Channel & i'm not either, my sister is a pole dancer and she's very high maintenance\\
    CTRL &  do you have a favorite color? mine is red \\
    CTRL Online & my sister is a pole dancer, and she's very high maintenance \\
    \hline \hline
    Grounding  & \linebreakcell{i love playing softball . my father is in the army . i recently was fired and had to get a new job . \\ my mom passed away when i was young . i work in the insurance business .
}
\\ 
    $u_T$  & my german shepherd is barnaby \\ 
    Ground Truth & did you name him after someone \\ \hline
    Direct Model & do you have any pets \\ \hdashline
    Reranking & what do you do for a living \\
    Online Decoding & that's a cute name \\
    CTRL & my mom passed away when i was young \\
    CTRL Online & i work in the insurance business and recently got fired \\
    \hline \hline
    Grounding & \linebreakcell{i read twenty books a year . i'm a stunt double as my second job .\\ i only eat kosher . i was raised in a single parent household .} \\
    $u_T$ & wow ! i do love a good horror movie . loving this cooler weather \\ 
    Ground Truth & but a good movie is always good . \\ \hline
    Direct model & i like the cooler weather too, but i prefer the hot weather. \\ \hdashline
    Reranking & what do you do for a living? i am a stunt double for a movie. \\
    Noisy Channel & i love the cooler weather too, i like to stay inside and read. \\ 
    CTRL & i have a second job as a stunt double. \\
    CTRL Online & i'm a stunt double as my second job. do you have any hobbies?\\ \hline \hline
    
Grounding & \linebreakcell{i've a german shepherd named barnaby . i am employed by the us postal service .\\ my father drove a car for nascar . my favorite color is gray .} \\
$u_T$ & i'm thirty two . i work in the banking industry , boring . \\
Ground truth &  i work for the us postal service . it is pretty boring as well . \\
    Direct Model & i work for the us postal service. \\
    Reranking: & that sounds boring. i work for the us postal service. \\
    Online Decoding & i work for the us postal service. do you have pets? \\
    CTRL & i work for the us postal service. \\
    CTRL Online & i am employed by the us postal service, boring too. \\ \hline \hline
    
Grounding & \linebreakcell{my family has raised horses for a generations . i dream of moving to the city .\\ i'm from a small town . i was raised on a horse farm . i want to be a lawyer .} \\
$u_T$ & marriage is fun . . we just bought our first house . \\
Ground truth &  we built a house on my family horse farm . we have over 5000 acres . \\
Direct Model &  where are you from? i'm from a small town in the midwest. \\ \hdashline
Reranking & that's great! where are you from? i'm from a small town. \\
Online Decoding & that's great! i grew up on a horse farm and have always wanted to move to the city. \\
CTRL &  where are you from? i'm from a small town. \\
CTRL Online & my family has been raising horses for generations. i want to move to the city someday. \\ \hline 
        
        \multicolumn{2}{c}{\textbf{DSTC9 Track1}} \\ \hline

Grounding & You may bring 2 domestic pets, they must be kept on a short lead, or carried in a basket size 85x60x60cm. \\
$u_T$ & So pets of any size can go on the train? \\
Ground Truth & Yes, but for pets in a basket, the basket must be no larger than 85x60x60cm. Would you like me to book the train? \\ \hline
Direct Model & Pets must be able to fit into a basket that's 85x60x60cm. Can I help you with anything else? \\ \hdashline
Reranking & Yes, you may bring your pet on board on a short leash or in a basket 85x60x60cm. Do you have any other questions? \\
Online Decoding & You can bring 2 domestic pets on a short leash or in a basket 85x60x60cm. Is there anything else I can help you with? \\
CTRL & Yes, you may bring your pet on board on a short leash or in a 85x60x60cm basket. Anything else I can do for you? \\
CTRL Online & You may bring 2 domestic pets, but they must be kept on a short lead or carried in a basket 85x60x60cm. (...) \\ \hline \hline

    Grounding & English is spoken at Taylor Hotel San Francisco. \\
    $u_T$ & Thanks, and do they speak any languages other than English? \\ 
    Ground Truth & Only English is spoken at the Taylor Hotel San Francisco. Would you like me to find any more details? \\ \hline
    Direct Model & Yes, the hotel does speak English. Would you like to make a reservation? \\ \hdashline
    Reranking & No, they only speak English at the hotel. Would you like me to book a room for you? \\
    Online Decoding & No, they only speak English at the hotel. Would you like me to book a room for you? \\
    CTRL & No, they only speak English at the hotel. Is there anything else I can help you with? \\
    CTRL Online & No, they only speak English at the hotel. Is there anything else I can help you with today? \\ \hline \hline
    
    Grounding & Yes it can be, as long as they have access to a payment card for identification. \\
    $u_T$ & Can my son pick up these tickets for me? \\
    Ground Truth & Yes as long as they have access to the payment card for identification. Is there anything else I can help you with? \\ \hline
    Direct Model & Yes, your son can pick up the tickets for you. Anything else I can do for you? \\ \hdashline
    Reranking & Yes as long as they have access to the payment card for identification. Is there anything else I can help you with? \\
    Online Decoding & Yes as long as they have access to the payment card for identification. Is there anything else I can help you with today? \\ 
    CTRL & Yes as long as he has access to the payment card for identification. Is there anything else I can help you with? \\
    CTRL Online & Yes as long as they have access to the payment card for identification. Is there anything else I can help you with? \\ \hline
 \end{tabular}}
\end{table*}

\newpage

\begin{table*}
\small
 \resizebox{\textwidth}{!}{\begin{tabular}{|l|l|}
    \multicolumn{2}{c}{\textbf{WoW}} \\ \hline
    Grounding & She was the founder and namesake of the Chanel brand. \\ 
    $u_T$: & - \\
Ground truth & I learned that the founder of the Chanel brand's first name is Coco. Are you familiar with the brand? \\ \hline
Direct Model & Have you ever heard of Coco Chanel? She was the founder of the Chanel brand. \\ \hdashline
Reranking: & Have you heard of Coco Chanel? She was the founder and namesake of the Chanel brand \\
Online Decoding: & I love Coco Chanel. She was the founder and namesake of the Chanel brand. \\
CTRL &  She was the founder and namesake of the Chanel brand. \\
CTRL Online & Coco Chanel was the founder and namesake of the Chanel brand \\\hline \hline

Grounding & \linebreakcell{The giant panda ("Ailuropoda melanoleuca", "black and white cat-foot"; , "big bear cat"),\\ also known as panda bear or simply panda, is a bear native to south central China.} \\
$u_T$ & - \\
Ground truth & Hi! I love giant pandas, they are my favourite animal. they're also called "big bear cat" sometimes? how about you? \\ \hline
Direct Model & Have you ever seen a giant panda? They are native to south central China. \\ \hdashline
Reranking & My favorite animal is the giant panda. They are native to south central China. \\
Online Decoding & Giant pandas are so cute! They are native to south central China. \\
CTRL &  The giant panda is a bear native to south central China. \\
CTRL Online & The giant panda is a bear native to south central China. \\ \hline

    \multicolumn{2}{c}{\textbf{Doc2Dial}} \\ \hline

Grounding & Financial difficulties \\
$u_T$: & what does my circumstance have to be to get a general forbearance? \\
Ground truth & Are you having financial difficulties? \\ \hline
Direct Model & Do you have financial difficulties? \\
Reranking & You would have to have financial difficulties \\
Online Decoding & Financial difficulties are one of the requirements to get a general forbearance \\
CTRL & Financial difficulties \\
CTRL Online & Financial difficulties \\ \hline \hline

Grounding & you may request a hearing online. \\
$u_T$ & Yes, I disagree with the determination you made at the reconsideration level. \\
Ground truth & You may then request a hearing online. \\ \hline
Direct Model & You can request a hearing online. \\ \hdashline
Reranking & You can request a hearing online if you disagree with the determination. \\
Online Decoding & In that case you may request a hearing online. \\
CTRL & You may request a hearing online. \\
CTRL Online & You may request a hearing online. \\ \hline

 \end{tabular}}
\end{table*}

\begin{table*}
\small
 \resizebox{\textwidth}{!}{\begin{tabular}{|l|l|}
 \hline
    Grounding & In computing, React (sometimes styled React.js or ReactJS) is a JavaScript library for building user interfaces. \\
    $u_T$ & I'm not sure I've heard of it before, what is it? \\ \hline
    $\frac{\lambda_1}{\lambda_2} = 0.5$ & It is a JavaScript library for building user interfaces. \\
    $\frac{\lambda_1}{\lambda_2} = 2.0$ & React (sometimes styled React.js or ReactJS) is a JavaScript library for building user interfaces. \\
    \hline \hline
    Grounding & David Suchet stars as the eponymous detective, Agatha Christie's fictional Hercule Poirot. \\
    $u_T$ & No do you know what's it about? \\
    $\frac{\lambda_1}{\lambda_2} = 0.5$ & It's about a detective named Hercule Poirot, played by David Suchet. \\
    $\frac{\lambda_1}{\lambda_2} = 2.0$ & David Suchet stars as the eponymous detective, Agatha Christie's fictional Hercule Poirot. \\ \hline \hline
    Grounding & my friends are all skateboarders . i've a broken arm . i am a skateboarder . my favourite skater is rob dryden . \\
    $u_T$ & oh no ! what is wrong with it ? are you a pro skateboarder \\
    $\frac{\lambda_1}{\lambda_2} = 0.5$ & i broke my arm trying to skateboard with my friend rob dryden \\
    $\frac{\lambda_1}{\lambda_2} = 1.4$ & i broke my arm and my friends are all skateboarders \\ \hline \hline
    Grounding & \linebreakcell{i like to work on vintage cars . i own two vintage mustangs . i like to go to country concerts on weekends\\ . my favorite music is country . i have two dogs .} \\
    $u_T$ & that's amazing . i'm currently studying spanish . \\
    $\frac{\lambda_1}{\lambda_2} = 0.5$ & i love to go to country concerts on the weekends \\
    $\frac{\lambda_1}{\lambda_2} = 1.4$ & i love to go to country concerts on the weekends with my dogs \\ \hline \hline
    Grounding & i don't drink soda . i've a boyfriend . i live in a house . i like to eat hamburgers and french fries . \\
    $u_T$ & i try to work and stay busy , i am in between jobs right now . \\
    $\frac{\lambda_1}{\lambda_2} = 0.5$ & what do you do for a living? \\
    $\frac{\lambda_1}{\lambda_2} = 1.4$ & do you like hamburgers and french fries? i love them. \\ \hline \hline
    Grounding & \linebreakcell{i volunteer my time with a local bunny rescue . my favorite color is navy blue \\. i work fulltime in a shipyard . i donate a part of my salary to charity each month .} \\
    $u_T$ & wow that's exciting good for you . i'm learning about computers , programming \\
    $\frac{\lambda_1}{\lambda_2} = 0.5$ & that's cool. i work at a shipyard. i love computers and programming. \\
    $\frac{\lambda_1}{\lambda_2} = 1.4$ & cool. i work in a shipyard full time. what do you do in your spare time?\\ \hline \hline
    Grounding & \linebreakcell{i'm a descendant of christopher columbus . i love to cook paella . i've a weakness for fish and chips\\ . i visit europe twice a year . i am an art major in college .} \\
    $u_T$ & europe must be very beautiful my sister went there \\
    $\frac{\lambda_1}{\lambda_2} = 0.5$ & it is. i love it. i've been there twice a year since i was in college. \\
    $\frac{\lambda_1}{\lambda_2} = 1.4$ & it is. i love it. i go twice a year, and i cook paella and fish and chips \\ \hline \hline
    Grounding & \linebreakcell{my family has raised horses for a generations . i dream of moving to the city\\ . i'm from a small town . i was raised on a horse farm . i want to be a lawyer .} \\
    $u_T$ & keep it in your family , that's nice for future generations \\
    $\frac{\lambda_1}{\lambda_2} = 0.5$ & thank you. i want to move to the city and live on a horse farm. \\
    $\frac{\lambda_1}{\lambda_2} = 1.4$ & thank you. i want to move to the city one day, but i live in a small town \\ \hline \hline
    
    Grounding & Opening hours are from 11am to 5pm on Wednesdays only. \\
    $u_T$ & What time does the Oakland Museum open? \\
    $\frac{\lambda_1}{\lambda_2} = 0.2$ & \linebreakcell{The museum opens at 11am on Wednesdays and at 5pm on Sundays.\\ Do you have any other questions about the museum?} \\
    $\frac{\lambda_1}{\lambda_2} = 1.4$ & The museum opens at 11am on Wednesdays only. Is there anything else I can help you with? \\ \hline\hline
    Grounding & \linebreakcell{Cathedral of Saint Mary of the Assumption is open from Monday to Friday from 9:30 am to 4:00 pm. \\ On Saturday it is open from 11:00 am to 5:30 pm and on Sunday it is open from 9:00 am to 3:30 pm.} \\
    $u_T$ &  Yeah, is it open on weekends too? \\
    $\frac{\lambda_1}{\lambda_2} = 0.2$ & \linebreakcell{The Cathedral of Saint Mary of the Assumption is open from Monday to Friday from 9:30 am to 4:00 pm.\\ Is there anything else I can help you with today?} \\
    $\frac{\lambda_1}{\lambda_2} = 1.4$ & \linebreakcell{It is open from Monday to Friday from 9:30 am to 4:00 pm. \\On Saturday it opens at 11:00 am to 5:30 pm and on Sunday it is open at 9:00am to 3:30.\\ Is there anything else I can help you with?} \\ \hline
 \end{tabular}}
\end{table*}

\end{document}